\documentclass[letterpaper, 10 pt, conference]{ieeeconf} 

\usepackage{mathtools}
\usepackage[listings,skins]{tcolorbox}
\usepackage{booktabs}
\usepackage[flushleft]{threeparttable}
\usepackage{amssymb}

\usepackage{float}
\usepackage{hyperref}
\usepackage{calc}

\usepackage{caption}
\usepackage{subcaption}

\usepackage{longtable}
\usepackage{array}
\usepackage{multirow}
\usepackage{multicol}

\IEEEoverridecommandlockouts        
\AtBeginEnvironment{thebibliography}{\footnotesize}

\usepackage{etoolbox}
\AtBeginEnvironment{longtable}{\small}
\definecolor{lightgray}{HTML}{F1F0F5}

\usepackage{booktabs}
\usepackage{rotating} 

\usepackage{multirow}
\usepackage{multicol}
\usepackage{amsthm}

\usepackage{colortbl}

\definecolor{gr}{rgb}{0.921, 0.972, 0.905}
\definecolor{pink}{rgb}{0.972, 0.905, 0.917}
\definecolor{redishh}{rgb}{0.9, 0.17, 0.31}
\definecolor{redish}{rgb}{1.0, 0.01, 0.24}
\definecolor{antique}{rgb}{0.57, 0.36, 0.51}
\definecolor{darkcandy}{rgb}{0.64, 0.0, 0.0}
\definecolor{pastel}{rgb}{0.09, 0.45, 0.27}

\IEEEoverridecommandlockouts        
\title{
Externally Validated Longitudinal GRU Model for Visit‑Level 180‑Day Mortality Risk in Metastatic Castration‑Resistant Prostate Cancer  
}

\author{
    \begin{tabular}{ccc}
        Javier Mencia-Ledo & Mohammad Noaeen & Zahra Shakeri \\
        \textit{University of Toronto} & \textit{University of Toronto} & \textit{University of Toronto} \\
        Toronto, Canada & Toronto, Canada & Toronto, Canada \\
        j.mencialedo@mail.utoronto.ca & m.noaeen@utoronto.ca & zahra.shakeri@utoronto.ca
    \end{tabular}
}

\begin{document}
\maketitle
\thispagestyle{empty}
\pagestyle{empty}

\begin{abstract} Metastatic castration-resistant prostate cancer (mCRPC) is a highly aggressive disease with poor prognosis and heterogeneous treatment response. In this work, we developed and externally validated a visit-level 180-day mortality risk model using longitudinal data from two Phase III cohorts (n=526 and n=640). Only visits with observable 180-day outcomes were labeled; right-censored cases were excluded from analysis. We compared five candidate architectures: Long Short-Term Memory, Gated Recurrent Unit (GRU), Cox Proportional Hazards, Random Survival Forest (RSF), and Logistic Regression. For each dataset, we selected the smallest risk-threshold that achieved an 85\% sensitivity floor. The GRU and RSF models showed high discrimination capabilities initially (C-index: 87\% for both). In external validation, the GRU obtained a higher calibration (slope: 0.93; intercept: 0.07) and achieved an PR-AUC of 0.87. Clinical impact analysis showed a median time‑in‑warning of 151.0 days for true positives (59.0 days for false positives) and 18.3 alerts per 100 patient-visits. Given late-stage frailty or cachexia and hemodynamic instability, permutation importance ranked BMI and systolic blood pressure as the strongest associations. These results suggest that longitudinal routine clinical markers can estimate short-horizon mortality risk in mCRPC and support proactive care planning over a multi-month window. \end{abstract}
\definecolor{amaranth}{rgb}{0.9, 0.17, 0.31}
\definecolor{gr}{rgb}{0.55, 0.71, 0.0}
\definecolor{ashgrey}{rgb}{0.7, 0.75, 0.71}


\section{INTRODUCTION}

Metastatic castration-resistant prostate cancer (mCRPC) has a poor prognosis and an unpredictable clinical course, making short-term mortality risk prediction an urgent but unsolved challenge \cite{Slootbeek2024}. Most patients develop bone metastases (nearly 90\%) and treatment responses vary widely \cite{Slootbeek2024}. As a result, clinicians have no definitive biomarkers to signal an imminent decline. Median survival is only about 2-3~years, and five-year survival is approximately 30\%-40\% \cite{Huo2025,Freedland2024,halabi2014updated}. The critical question is how to detect when a patient transitions into a short-horizon high-risk state that warrants proactive care planning.

Existing prognostic models (e.g., Cox proportional hazards, random survival forests, logistic regression) can achieve strong predictive performance \cite{Huang2023,Price2018,Wang2019}. However, these approaches are typically static and provide risk estimates only at a single landmark time point rather than updating risk at each new visit \cite{Huang2023,Price2018,Wang2019}. Recent deep learning models capture longitudinal patient deterioration, but they often omit early-warning indicators and rarely undergo external validation on independent cohorts \cite{Reis2025,Kaddes2025,Khan2025,Thorsen-Meyer2022}.

To address these gaps, we developed an externally validated visit-level model that predicts 180-day mortality risk at each clinic encounter in mCRPC. The model uses a Gated Recurrent Unit (GRU) architecture and relies only on routinely collected structured clinical data (e.g., vital signs and functional status). The approach requires no new assays, imaging, or specialized infrastructure, which supports integration into existing clinical workflows. Our work makes three contributions to prognostic modeling in mCRPC. First, our model generates time-updated risk scores at each visit rather than a single static prediction. Second, our work quantifies the clinical utility of an alert state using two metrics. Alert density is the number of alerts per 100 patient-visits. Time-in-warning is the number of days a patient remains in an alert state before death. Third, external validation in an independent Phase~III trial with a different experimental agent supports generalizability under distribution shift.

The GRU model showed strong performance in both the development and validation cohorts. The model achieved a concordance index of approximately 0.87 in the development cohort and an AUC-ROC of 0.89 in the external trial. The risk predictions were well calibrated, with a calibration slope of approximately 0.93. At 85\% sensitivity, the model provided a median lead time of 151~days before mortality events. The alert burden remained modest, with approximately 18 alerts per 100 patient-visits in the external cohort. These results indicate that routinely collected longitudinal markers can produce a robust short-term mortality warning signal that supports proactive care planning in late-stage cancer.


\section{Methods}
\subsection{Problem Formalization}
Each patient $i$ is represented by a sequence of clinical visits $V_i = \{v_{i,1}, \dots, v_{i,T_i}\}$, where each visit $v_{i,t}$ consists of a feature vector $\mathbf{x}_{i,t} \in \mathbb{R}^d$ containing physiological and temporal markers. The model architecture processes these inputs to generate a representation $\mathbf{Z}_{i,t} = f_{Model}(\mathbf{x}_{i,1}, \dots, \mathbf{x}_{i,t})$ after every visit, encoding the patient’s cumulative physiological trajectory to produce a time-updated risk score $S(\mathbf{x}_{i,t}) \in [0, 1]$. We define a visit-level label $y_{i,t} = 1$ if the recorded death date, $T_i$, satisfies $0 < T_i - t \leq 180$ days, and a visit is labeled $y_{i,t} = 0$ only when the patient is known to be alive for at least 180 days after $t$. Hence, $S(\mathbf{x}_{i,t})$ score represents the probability of death within a 180-day \textit{look-ahead} window after the indexed visit, i.e. $\text{p}(y_{i,t} = 1)$. Visits without complete 180-day outcome ascertainment (e.g., administrative end of follow-up before 180 days without recorded death) are treated as right-censored and are not used for supervised loss or metric computation. In this setup, a \textquotesingle{}negative\textquotesingle{} example represents a visit followed by at least 180 days of confirmed survival (i.e., the patient remained under observation and alive through day 180), rather than the absence of a recorded death within the dataset. We define the alert state $A_{i,t}$ as a binary indicator triggered when the risk score exceeds an optimized threshold $\theta$, formulated as $A_{i,t} = \mathbf{1}(S(\mathbf{x}_{i,t}) \geq \theta)$, where $\theta$ is calibrated to maintain an 85\% sensitivity floor for deaths occurring within the 180-day window for each CV fold and the external cohort, respectively. To capture a clinical safety constraint, we report performance at a fixed operating point. This approach separates model ranking and calibration from arbitrary threshold selection and supports fair comparisons across architectures. Finally, we evaluate clinical utility using two primary operational metrics: 
\begin{itemize}
    \item  Alert Density, defined as alerts per 100 patient-visits $\left( \frac{\sum A_{i,t}}{\sum 1} \times 100 \right)$, to quantify alarm burden.
    \item Time-in-Warning ($\tau$), representing the total duration a true positive case remains in an active alert state prior to the event ($\tau_i = \sum_{t=1}^{T_i} (t_{v, i, t} - t_{v, i, t-1}) \cdot A_{i,t}$)
\end{itemize}

\begin{table*}[ht]
\centering
\caption{Demographic and Clinical Characteristics Stratified by Survival Status (Development vs. External Validation Cohort)}
\label{tab:data}
\footnotesize
\setlength{\tabcolsep}{4pt}
\begin{tabular}{@{} l ccc c >{\columncolor{lightgray}}c >{\columncolor{lightgray}}c >{\columncolor{lightgray}}c @{}}
\toprule

& \multicolumn{3}{c}{\textbf{Model Development Cohort ($n=526$)}} & & \multicolumn{3}{>{\columncolor{lightgray}}c@{}}{\textbf{External Validation Cohort ($n=640$)}} \\
\cmidrule{2-4} \cmidrule{6-8}
\textbf{Characteristic} & \textbf{Overall} & \textbf{Alive} & \textbf{Dead} & & \textbf{Overall} & \textbf{Alive} & \textbf{Dead} \\
& ($n=526$) & ($n=450$) & ($n=76$) & & ($n=640$) & ($n=210$) & ($n=430$) \\
\midrule

\textbf{Sex} & & & & & & & \\
\hspace{5pt}Male & 526 (100\%) & 450 (100\%) & 76 (100\%) & & 640 (100\%) & 210 (100\%) & 430 (100\%) \\
\addlinespace[3pt]

\textbf{Race} & & & & & & & \\
\hspace{5pt}Asian / Oriental & 13 (2.5\%) & 13 (2.9\%) & 0 (0\%) & & 36 (5.6\%) & 11 (5.2\%) & 25 (5.8\%) \\
\hspace{5pt}Black & 25 (4.8\%) & 21 (4.7\%) & 4 (5.3\%) & & 17 (2.7\%) & 5 (2.4\%) & 12 (2.8\%) \\
\hspace{5pt}White & 433 (82.3\%) & 370 (82.2\%) & 63 (82.9\%) & & 580 (90.6\%) & 190 (90.5\%) & 390 (90.7\%) \\
\hspace{5pt}Other & -- & -- & -- & & 7 (1.1\%) & 4 (1.9\%) & 3 (0.7\%) \\
\hspace{5pt}Missing / NA & 55 (10.5\%) & 46 (10.2\%) & 9 (11.8\%) & & -- & -- & -- \\
\addlinespace[3pt]

\textbf{Region} & & & & & & & \\
\hspace{5pt}E. Europe & 84 (16\%) & 78 (17.3\%) & 6 (7.9\%) & & 131 (20.5\%) & 33 (15.7\%) & 98 (22.8\%) \\
\hspace{5pt}N. America & 139 (26.4\%) & 112 (24.9\%) & 27 (35.5\%) & & 87 (13.6\%) & 31 (14.8\%) & 56 (13.0\%) \\
\hspace{5pt}S. America & -- & -- & -- & & 95 (14.8\%) & 38 (18.1\%) & 57 (13.3\%) \\
\hspace{5pt}W. Europe & 247 (47\%) & 213 (47.3\%) & 34 (44.7\%) & & 229 (35.8\%) & 75 (35.7\%) & 154 (35.8\%) \\
\hspace{5pt}Other & 50 (9.5\%) & 43 (9.6\%) & 7 (9.2\%) & & 98 (15.3\%) & 33 (15.7\%) & 65 (15.1\%) \\
\hspace{5pt}Missing / NA & 6 (1.1\%) & 4 (0.9\%) & 2 (2.6\%) & & -- & -- & -- \\
\addlinespace[3pt]

\textbf{Age Baseline} & & & & & & & \\
\hspace{5pt}40-49 & 9 (1.7\%) & 9 (2\%) & 0 (0\%) & & 7 (1.1\%) & 2 (1.0\%) & 5 (1.2\%) \\
\hspace{5pt}50-59 & 87 (16.5\%) & 76 (16.9\%) & 11 (14.5\%) & & 95 (14.8\%) & 26 (12.4\%) & 69 (16.0\%) \\
\hspace{5pt}60-69 & 225 (42.8\%) & 194 (43.1\%) & 31 (40.8\%) & & 270 (42.2\%) & 90 (42.9\%) & 180 (41.9\%) \\
\hspace{5pt}70+ & 205 (39\%) & 171 (38\%) & 34 (44.7\%) & & 268 (41.9\%) & 92 (43.8\%) & 176 (40.9\%) \\
\addlinespace[3pt]

\textbf{ECOG Baseline} & & & & & & & \\
\hspace{5pt}0 & 313 (59.5\%) & 266 (59.1\%) & 47 (61.8\%) & & 300 (46.9\%) & 129 (61.4\%) & 171 (39.8\%) \\
\hspace{5pt}1 & 202 (38.4\%) & 175 (38.9\%) & 27 (35.5\%) & & 311 (48.6\%) & 79 (37.6\%) & 232 (54.0\%) \\
\hspace{5pt}2 & 8 (1.5\%) & 6 (1.3\%) & 2 (2.6\%) & & 29 (4.5\%) & 2 (1.0\%) & 27 (6.3\%) \\
\addlinespace[3pt]

\textbf{Trial Drug: Lenalidomide} & & & & & & & \\
\hspace{5pt}No & 419 (79.7\%) & 359 (79.8\%) & 60 (78.9\%) & & -- & -- & -- \\
\hspace{5pt}Yes & 103 (19.6\%) & 87 (19.3\%) & 16 (21.1\%) & & -- & -- & -- \\
\hspace{5pt}NA & 4 (0.8\%) & 4 (0.9\%) & 0 (0\%) & & -- & -- & -- \\
\addlinespace[3pt]

\textbf{Trial Drug: Aflibercept} & & & & & & & \\
\hspace{5pt}No & -- & -- & -- & & 640 (100\%) & 210 (100\%) & 430 (100\%) \\
\hspace{5pt}Yes & -- & -- & -- & & 0 (0\%) & 0 (0\%) & 0 (0\%) \\
\addlinespace[3pt]

\textbf{Continuous Variables} & \multicolumn{3}{c}{Mean (SD)} & & \multicolumn{3}{>{\columncolor{lightgray}}c@{}}{Mean (SD)} \\
\hspace{5pt}Total Cycles & 9.2 (5.2) & 9.6 (5.2) & 6.7 (4.1) & & 10.02 (6.94) & 13.66 (9.15) & 8.61 (5.23) \\
\hspace{5pt}Number of Visits & 15.4 (6.9) & 15.7 (7.0) & 14.1 (6.3) & & 11 (7.8) & 14.8 (9.7) & 9.1 (5.8) \\
\hspace{5pt}Follow-Up Length (Days) & 286 (146.4) & 283.5 (149.8) & 300.7 (123.9) & & 200 (155.7) & 270 (203.18) & 165 (111.57) \\
\hspace{5pt}Range & 11--750 & 11--750 & 98--624 & & 1--1029 & 1--1029 & 1--918 \\

\bottomrule
\end{tabular}
\end{table*}

\subsection{Model Development}
\subsubsection{Overview}

All computational analyses and model developments were performed using Python version 3.12.1 and the pandas and TensorFlow libraries (versions 2.3.3 and 2.16.1, respectively)\footnote{The source code for the model is publicly available on \href{https://github.com/HIVE-UofT/mCRPC180DayMortality/blob/main/180DaySurvivalModel.ipynb}{GitHub}. The process of accessing the data will be shared upon request.}. To select the final model, five candidate models were compared, two neural networks, a Long Short Term Memory (LSTM) model , and a survival Gated Recurrent Unit (GRU) model \cite{Hochreiter1997, chung2014empiricalevaluationgatedrecurrent}, and three static models, Cox proportional hazards, random survival forests, and logistic regression models. All models were set to have a 85\% sensitivity floor and compared based on C-index, specificity, PPV, and NPV. 

We used stratified 3-fold cross-validation at the patient level to maintain sufficient sample size in the testing partitions for stable metric evaluation prior to external validation. To handle irregular sampling and missing information, we employed neural autoencoders to learn imputation patterns and reconstruct input within each training fold \cite{che2018recurrent}. Specifically, we used temporal autoencoders for the neural network models and denoising autoencoders for the static models \cite{vincent2008extracting}. To strictly prevent information leakage, the imputation weights and scaling parameters were derived solely from the training folds and all preprocessing models (i.e., scalers and autoencoders) were fit only within the training folds. Sequence models respect temporal order, but sequence-level imputation can still leak information if future steps reconstruct earlier missing values. For this reason, we treated imputation as a practical strategy for missing-data mitigation and identify strictly causal imputation as a key extension. Furthermore, for the denoising autoencoder, we restricted the input to data from previous visits only, ensuring no future information influenced the reconstruction of current features. Continuous features were standardized into $z$-scores to ensure unit variance across the feature space. To address class imbalance in the development cohort (14.4\% mortality), we used balanced batching (batch size 32). The external validation cohort had a higher mortality fraction (430/640, 67.2\%), which created a stringent distribution shift test for calibration and clinical alert burden. 

\subsubsection{Sequential Deep Learning Models}
Sequential architectures were implemented using both Long Short-Term Memory (LSTM) and Gated Recurrent Unit (GRU) networks to capture complex temporal dependencies within clinical visit sequences. These models utilized a stacked configuration consisting of two primary recurrent layers (64 and 32 units, respectively) integrated with batch normalization and dropout rates (0.3 and 0.2) to ensure regularization and training stability.

Input data were structured into fixed-length sequences representing up to 12 longitudinal clinical visits. For both architectures, the temporal autoencoder with a symmetric encoder-decoder structure (64-32-32-64 units) was employed for feature imputation. This pipeline was trained via a masked Mean Squared Error (MSE) loss function, which computed the reconstruction error exclusively on observed values to prevent bias from padded or missing entries. All sequential models were optimized over 25 epochs using the \texttt{Adam} optimizer, without early stopping. The final output layer employed a Time-Distributed Dense layer with a \texttt{sigmoid} activation function to generate a visit-specific mortality risk probability in the temporal sequence. Because the GRU/LSTM are unidirectional, the visit-level score at time $t$ is a function of $(x_{i,1}, \ldots, x_{i,t})$ only and later visits do not influence earlier visit predictions.

\subsubsection{Static Models}
For baseline comparison, three static models were implemented using landmarking at each patient visit \cite{van2008dynamic}. To calculate risk scores for the Logistic Regression (LR) model, a logit transformation was applied to a linear combination of features, utilizing an Elastic Net penalty (mixing $L1$ and $L2$ regularization) to manage feature collinearity while maintaining interpretability. The Cox Proportional Hazards (CoxPH) model calculated risk as the complement of the estimated survival probability at the 180-day horizon ($1 - \hat{S}(t=180 | X)$), assuming proportional hazards across the clinical features. Finally, the Random Survival Forest (RSF) \cite{ishwaran2008random} calculated risk scores by averaging the cumulative hazard functions across an ensemble of 200 de-correlated decision trees, utilizing log-rank splitting criteria to manage high-dimensional interactions between features.

\subsubsection{Clinical Impact Metrics} 
The clinical relevance of the predictive pipeline was scrutinized through a suite of operational metrics designed to simulate real-world implementation. These included the frequency of alerts, measured as alerts per 100 patient-visits, alongside the overall proportion of visits in alert state to quantify potential alarm fatigue. Temporal utility was measured by the median lead time, representing the duration between the first true positive alert and the mortality event, and the Time-in-Warning (TIW) for both true and false positive cases. 

Moreover, to identify the main contributors to model risk, we implemented permutation feature importance . This method permuted each feature across the batch and kept the original sequence length to quantify changes in predictive accuracy measured through average decreases in Concordance index (C-index) score per variable.

\section{Experimental Evaluation}
\subsection{Dataset and Data Preparation}
This study employs two distinct mCRPC Phase III clinical trial cohorts from the Project Data Sphere portal \cite{ProjectDataSphere}. The training and internal validation cohort (CC-5013-PC-002) included 526 patients, while the external validation cohort (EFC6546, control arm only) consisted of 936 patients, but 296 failed the screening and the final sample was reduced to 640. Outcomes were identified using the trial-recorded death date when available and the last known follow-up date otherwise. To avoid ambiguous negatives caused by right-censoring, visit-level labels were only assigned when the 180‑day look-ahead outcome was observable: death within 180 days (positive) or confirmed survival through 180 days (negative). Visits lacking 180‑day follow-up without a documented death were treated as right-censored and excluded from supervised training and evaluation.

To build the longitudinal predictive models, we selected eight clinical and temporal variables identified through literature review: current age, days since last visit, drug dosage reduction status, ECOG performance status, serious adverse event count, systolic blood pressure, body mass index (BMI), and pulse. Other variables that may have revealed a patient\textquotesingle{}s proximity to death, such as cycle number or drug interrupted status, were intentionally excluded. This yielded an EPV ratio of 9.5 and to address potential overfitting given this borderline EPV in the context of LSTM architecture, we implemented rigorous regularization including dropout (0.3), L2 regularization ($\lambda$ = 0.001), and 3-fold cross-validation with patient-level stratification. Nonetheless, within the external validation cohort, there were 430 events out of 640 total observations, resulting in an events-per-variable (EPV) of 53.8 for this subset.

The demographic and clinical profiles for both the model development cohort ($n=526$) and the external validation cohort ($n=640$) are detailed in Table~\ref{tab:data}. Both cohorts were entirely male, with most patients aged 60 or older (81.8\% development, 84.3\% validation) and predominantly White (82.3\% development, 90.6\% validation). While Western Europe was the most represented region in both groups, the external validation cohort included additional geographic diversity from South America (14.8\%). Mortality occurred in 14.4\% of the development cohort and 67.2\% of the external validation cohort during the study period. In both cohorts, patients who died completed fewer treatment cycles. Follow-up duration was markedly shorter among deceased patients in the external validation cohort (i.e., 165 vs 270 days), whereas follow-up duration was similar in the development cohort (i.e., 283.5 days alive vs 300.7 days deceased), consistent with differences in study observation structure.

\subsection{Model Performance}
\subsubsection{Mortality Prediction}
Table \ref{tab:model_performance} summarizes the predictive performance of the five candidate architectures, evaluated at a fixed sensitivity floor of 85\% to ensure clinical relevance across all comparisons. To maximise the other metrics, all models naturally targeted this threshold. The RSF and GRU architectures achieved the highest discriminative ability, both yielding a C-Index of 0.87. However, the GRU model demonstrated superior performance in identifying true negatives and minimizing false alarms, achieving the highest specificity (0.90) and balanced accuracy (0.88) among all tested models. Furthermore, the GRU showed significant gains in precision, with a PPV of 0.50, nearly double that of the RSF (0.27), while maintaining high clinical confidence in low-risk classifications with an NPV of 0.98. Notably, to achieve the 85\% sensitivity target, the static models (CoxPH, LR, and RSF) required significantly lower risk-score cutoffs (ranging from 0.12 to 0.27) compared to the higher operational thresholds utilized by the sequential GRU (0.62) and LSTM (0.53) architectures.

\begin{table*}[ht]
\centering
\footnotesize
\caption{Model Performance Comparison on Survival Prediction. To prevent data leakage, we performed all cross-validation splits at the patient level, with all visits from each patient kept within one fold. This table shows fold-to-fold variation under this patient-level partitioning.}
\label{tab:model_performance}
\resizebox{.8\linewidth}{!}{
\begin{tabular}{@{} l c c c c c c c @{}}
\toprule
\textbf{Model} & \textbf{C-Index} & \textbf{Sensitivity} & \textbf{Specificity} & \textbf{Bal. Acc} & \textbf{PPV} & \textbf{NPV} & \textbf{Threshold} \\
\midrule
CoxPH & 0.68 (0.03) & \textbf{0.85 (0.04)} & 0.37 (0.02) & 0.61 (0.03) & 0.14 (0.02) & 0.96 (0.02) & 0.23 \\
LR    & 0.70 (0.04) & \textbf{0.85 (0.01)} & 0.39 (0.02) & 0.63 (0.02) & 0.15 (0.02) & 0.96 (0.01) & 0.27 \\
RSF   & \textbf{0.87 (0.01)} & \textbf{0.85 (0.01)} & 0.72 (0.01) & 0.80 (0.01) & 0.27 (0.01) & \textbf{0.98 (0.01)} & 0.12\\
GRU   & \textbf{0.87 (0.02)}  & \textbf{0.85 (0.03)} & \textbf{0.90 (0.12)} & \textbf{0.88 (0.05)} & \textbf{0.50 (0.02)} & \textbf{0.98 (0.04)} & 0.62\\
LSTM  & 0.84 (0.05) & \textbf{0.85 (0.01)} & 0.86 (0.01) & 0.86 (0.01) & 0.42 (0.03) & 0.97 (0.02) & 0.53 \\

\bottomrule
\end{tabular}}
\begin{flushleft}
\footnotesize The table shows the performance of the candidate models Data presented as Mean (SD). RSF: Random Survival Forest; LR: Logistic Regression; GRU: Gated Recurrent Units; CoxPH: Cox Proportional Hazards; LSTM: Long Short Term Memory model. The threshold is the risk score cutoff to obtain 85\% sensitivity.
\end{flushleft}
\end{table*}

\begin{figure}[t]
\centering
\includegraphics[width=0.9\linewidth]{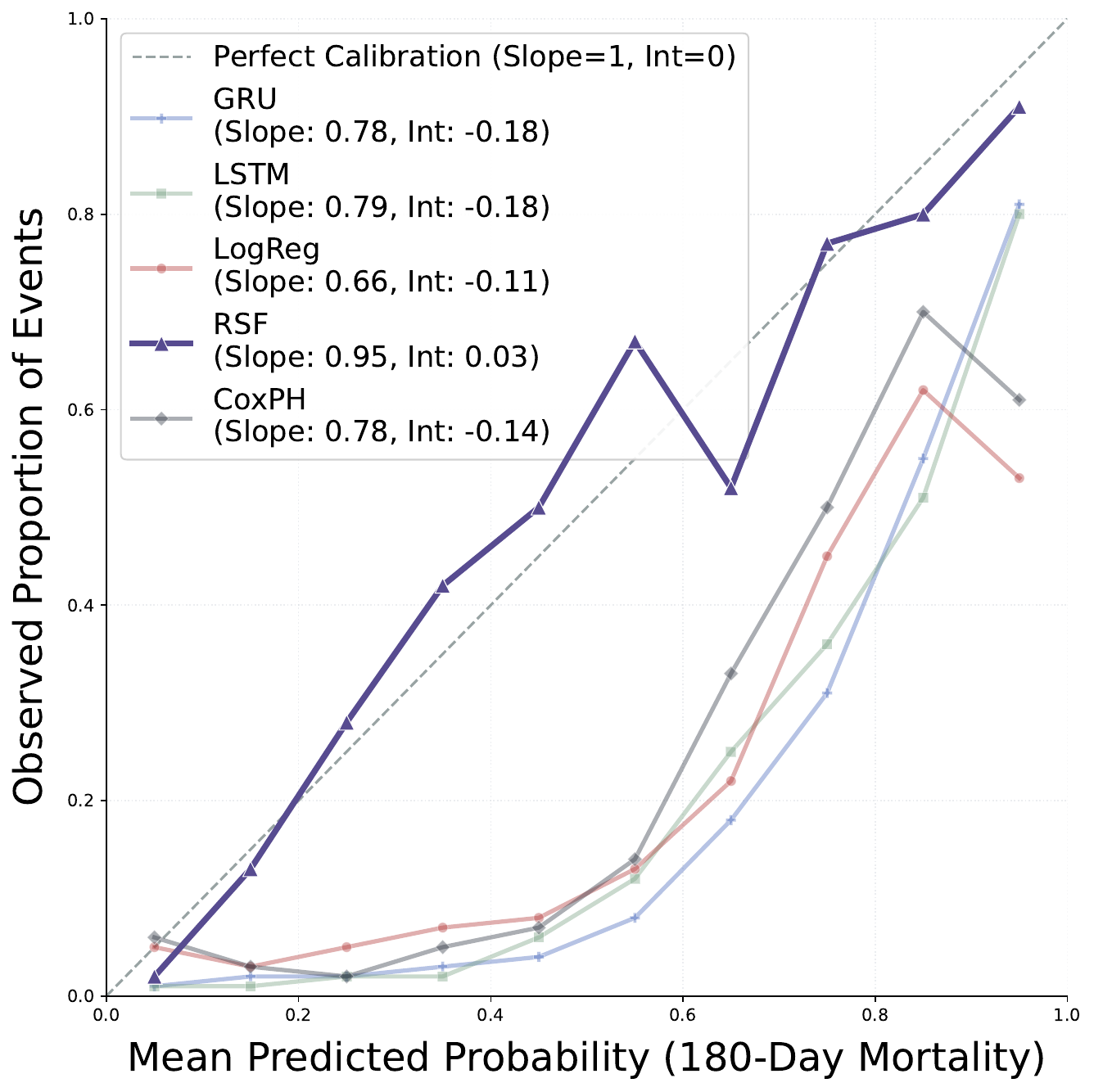}

\caption{\textbf{Comparison of Model Calibration for 180-Day Mortality Risk on the Development Cohort.}
The reliability diagram illustrates the predictive accuracy of the candidate architectures  at the visit level. This plots the mean predicted mortality risk against the observed proportion of death events. Perfect calibration is shown as the diagonal line.}
\label{fig:riskcurves}
\end{figure}

We evaluated calibration with reliability diagrams and summarized it using the intercept and slope from a linear fit \cite{van2019calibration}.
The fit related the observed event frequency to the mean predicted risk within deciles of predicted probability, with an ideal intercept of 0 and an ideal slope of 1. The overlaid calibration line plots for the candidate models are illustrated in Figure \ref{fig:riskcurves}. Across the development cohort, all models demonstrate an ability to correctly identify the extremes of the risk spectrum, with observed mortality rates aligning closely at both the lowest and highest predicted probability deciles. However, the RSF significantly outperforms the other architectures in the middle range of predicted probabilities, maintaining a slope of 0.95 and an intercept of 0.03 as it closely tracks the 45-degree ideal dashed line. In contrast, the sequential models (GRU and LSTM) and static baselines (LR and CoxPH) exhibit a notable deviation, characterized by a distinct underestimation of risk within the moderate probability range (0.3 to 0.6) before re-aligning at the highest risk deciles.

\subsubsection{External Validation}

\begin{figure*}[htbp]
    \centering
    \begin{subfigure}[b]{0.48\linewidth}
        \centering
        \includegraphics[width=\linewidth]{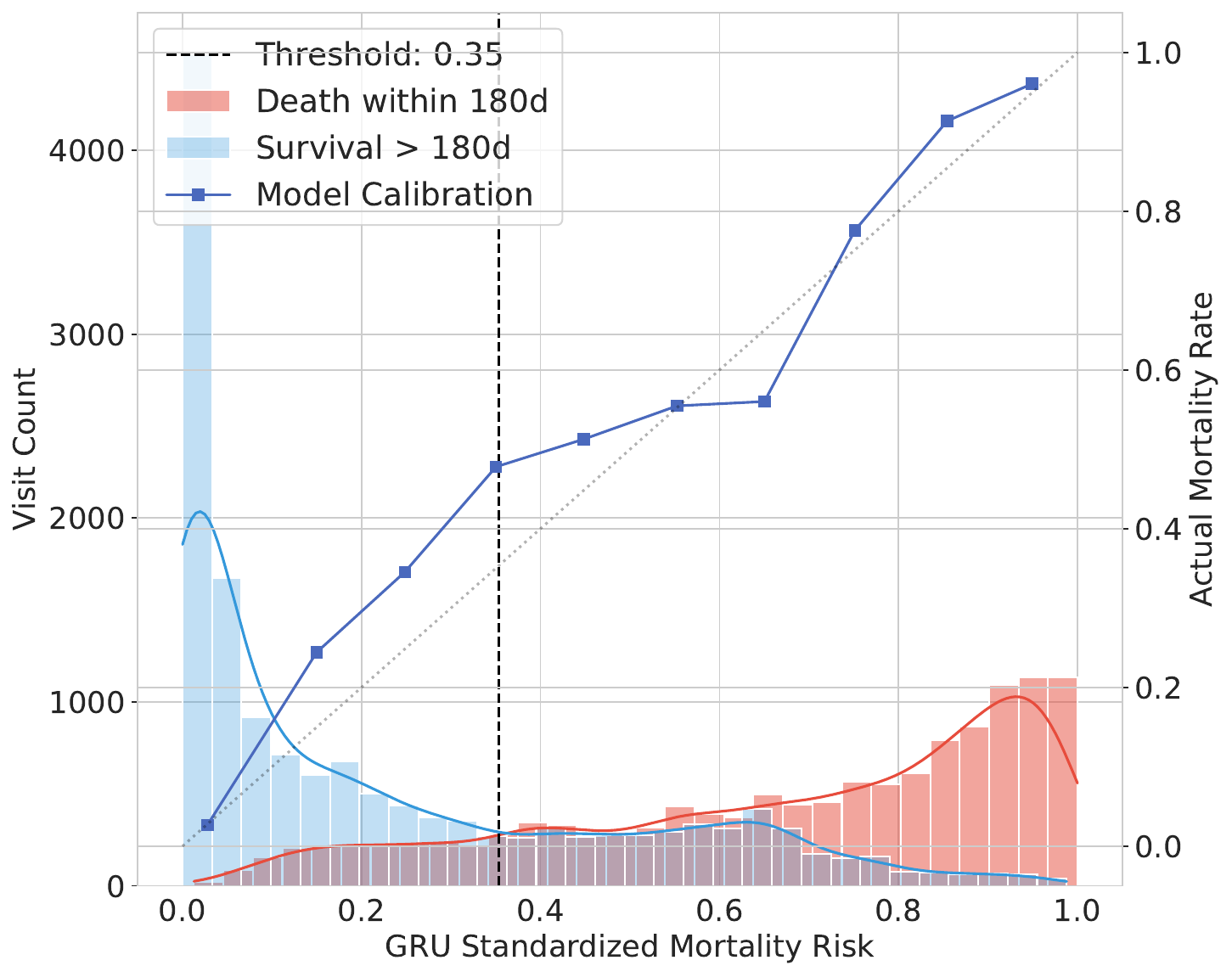} 
        \caption{GRU Model}
        \label{fig:first_sub}
    \end{subfigure}
    \hfill 
    \begin{subfigure}[b]{0.48\linewidth} 
        \centering
        \includegraphics[width=\linewidth]{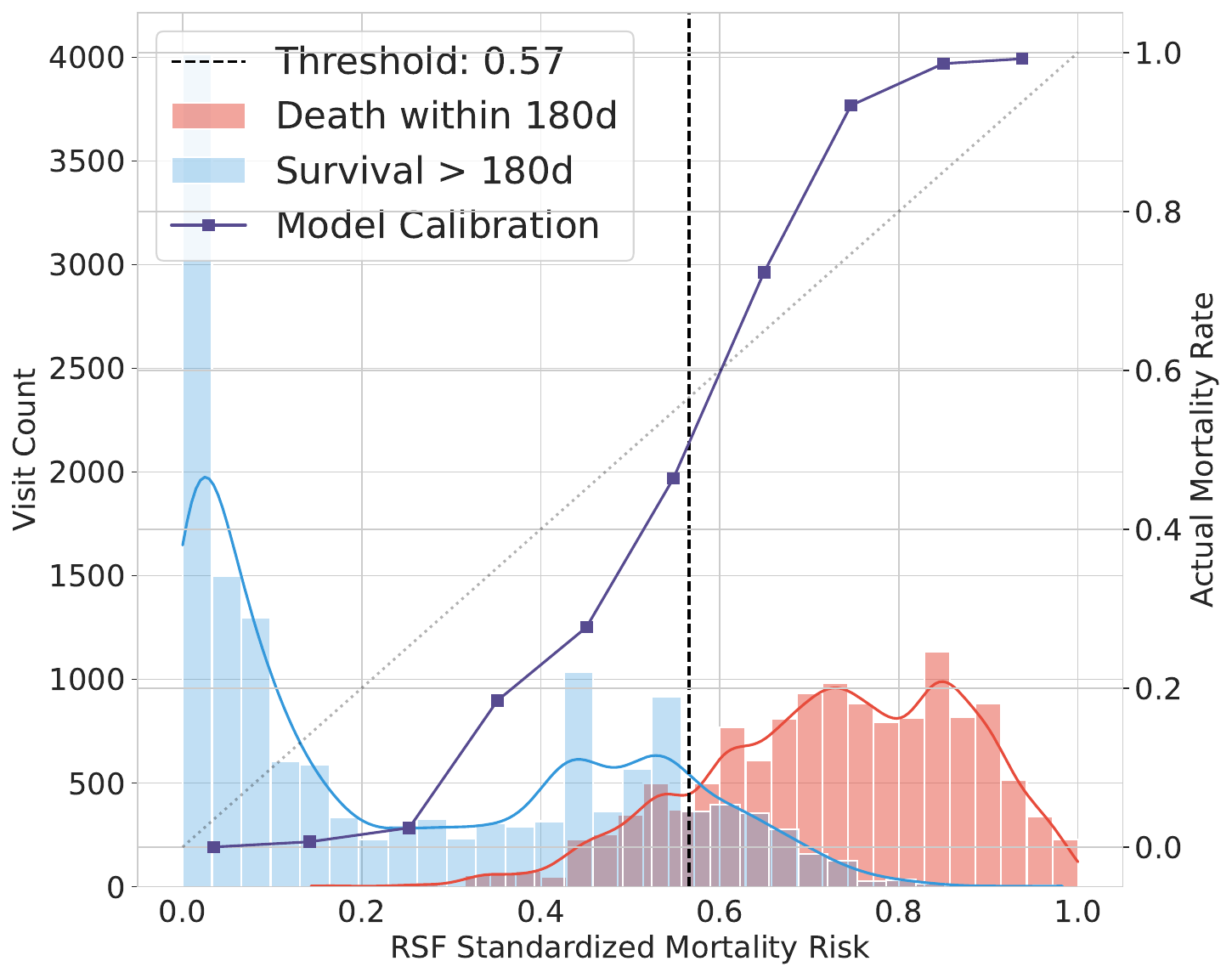}
        \caption{RSF Model}
        \label{fig:second_sub}
    \end{subfigure}
    
    \caption{\textbf{Mortality Risk Calibration and Distribution for the External Validation Cohort.} Both models are evaluated on their ability to predict death within a 180-day look-ahead window after each index visit date. Thresholds are optimized to prioritize high sensitivity for clinical alerting, with the GRU model leveraging temporal markers to update risk scores at each subsequent clinical visit.}
    \label{fig:hists}
\end{figure*}

Given that the Gated Recurrent Units (GRU) and the Random Survival Forest (RSF) models performed best in the development cohort, their performance was compared using the validation cohort. Here, the GRU improved its calibration slope to 0.93 and its intercept to 0.07 ($R^2 = 0.956$), getting remarkably close to the ideal targets. While the RSF achieved higher raw discrimination metrics, including a C-Index of 0.86 and a balanced accuracy of 0.87, it suffered from significant miscalibration, evidenced by a slope of 1.34 and an intercept of -0.20. Although the RSF showed higher specificity (0.89) and PPV (0.87) than the GRU’s 0.73 and 0.74, the GRU’s ability to provide well-calibrated risk estimates ensures that it does not systematically over- or under-estimate patient mortality risk. The histogram of the final risk scores is shown in Figure \ref{fig:hists}, which shows how the high separation in the extremes for the RSF gives rise to the miscalibrated calibration slope. On the other hand, the GRU plot shows a closer fit to the ideal calibration slope.

\begin{figure*}[t]
\centering
\subfloat[AUC with 95\% CI]{
  \includegraphics[width=0.23\textwidth]{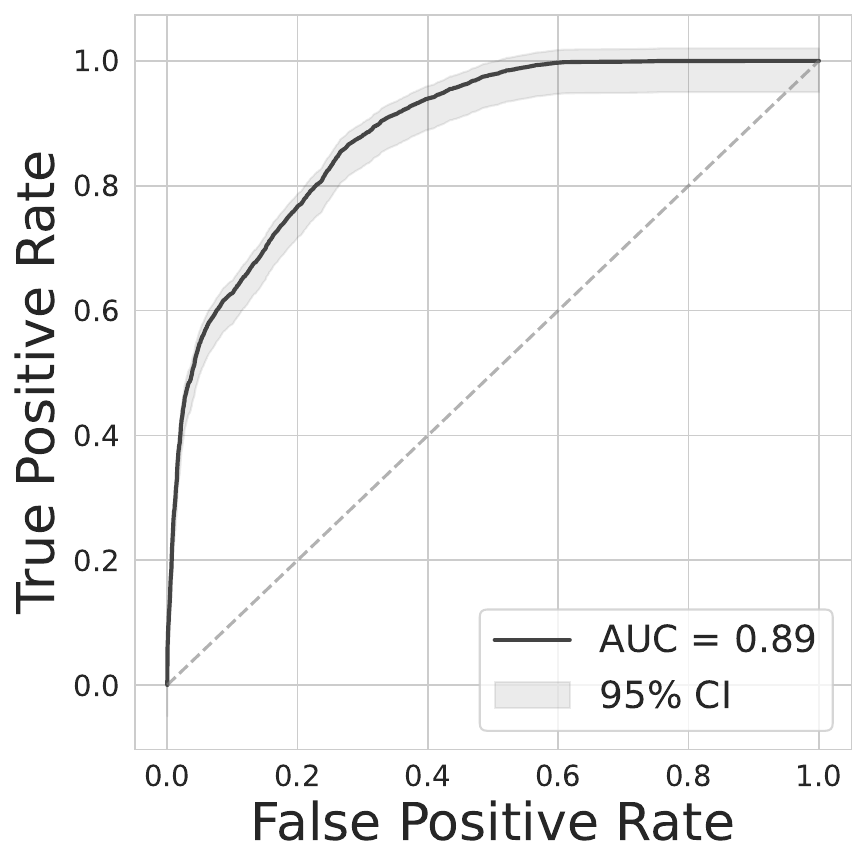}
  \label{fig:fig2_roc}
}
\subfloat[Precision--Recall curve ]{
  \includegraphics[width=0.23\textwidth]{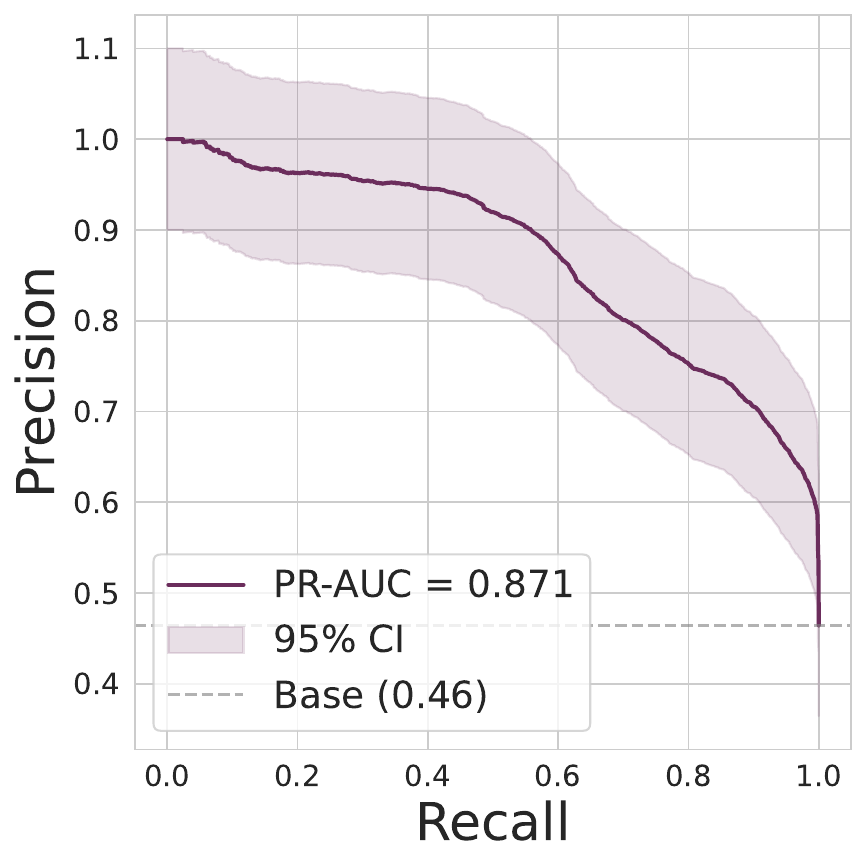}
  \label{fig:fig2_pr}
}
\subfloat[Sensitivity--specificity trade-off]{
  \includegraphics[width=0.23\textwidth]{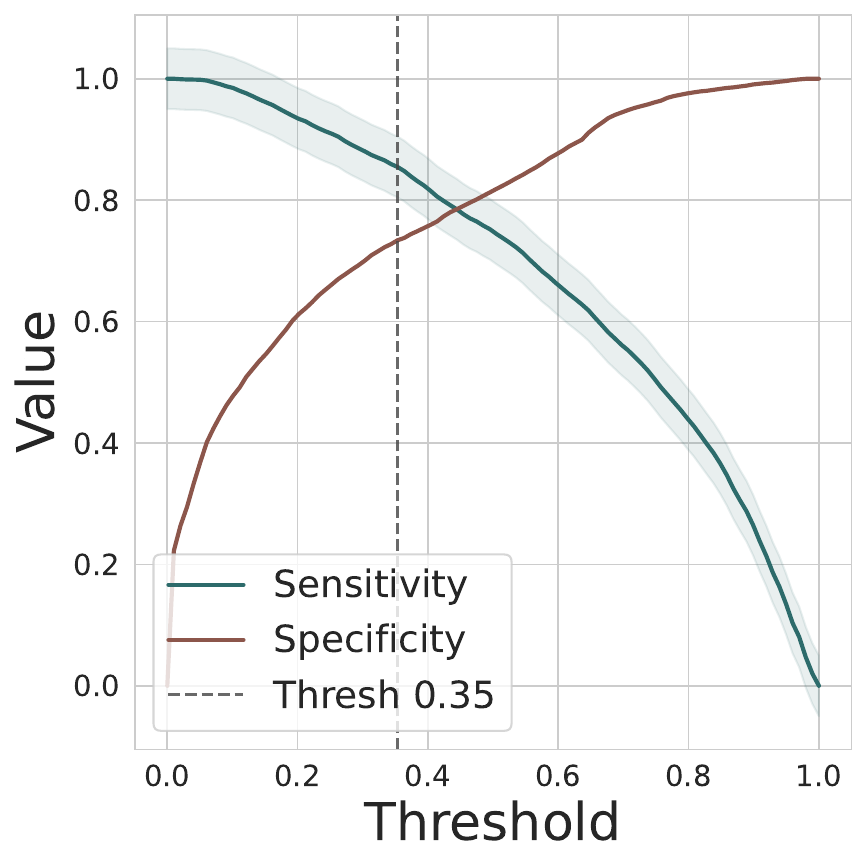}
  \label{fig:fig2_sensspec}
}
\subfloat[PPV--NPV trade-off]{
  \includegraphics[width=0.23\textwidth]{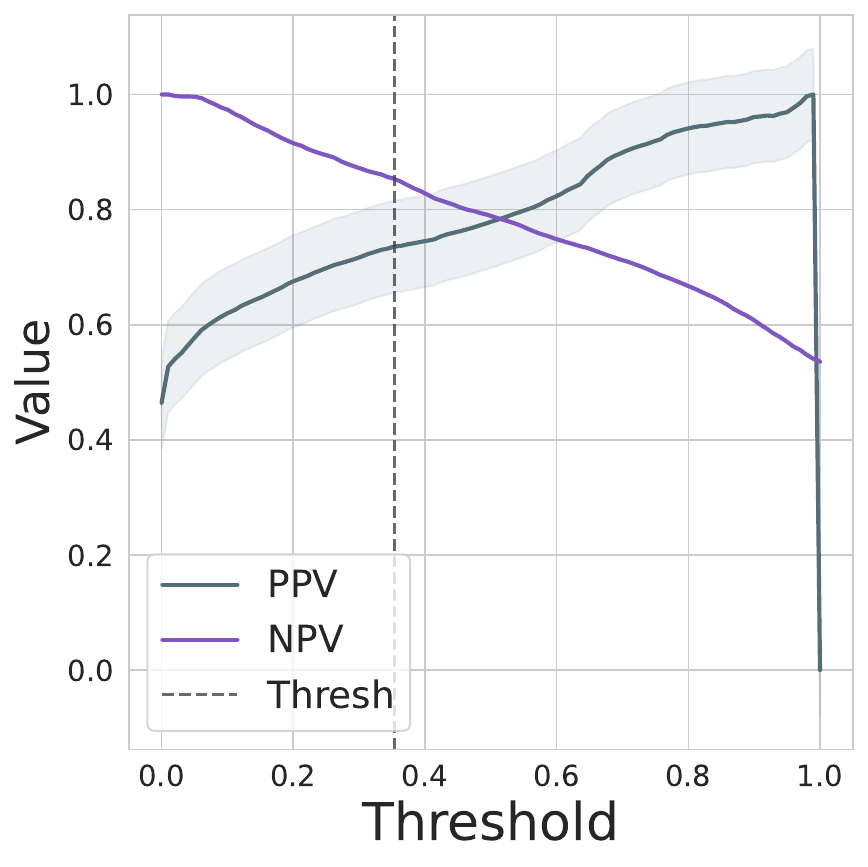}
  \label{fig:fig2_ppvnpv}
}
\caption{\textbf{Comprehensive Operating-Characteristic Evaluation of the GRU Risk Model on the External Validation Cohort at the Visit Level.} (a) ROC curve with 95\% confidence interval. To account for within-patient correlation in visit-level evaluation, we estimated confidence bands via patient-clustered nonparametric resampling, with details provided in the code. (b) Precision--Recall curve highlighting performance under class imbalance relative to the baseline event rate. (c) Sensitivity and specificity as a function of risk-score threshold, with the chosen operating threshold $\theta$ selected to satisfy the predefined sensitivity constraint. (d) PPV and NPV as a function of threshold, showing the resulting false-alarm versus missed-event trade-off at the selected 85\% sensitivity target.}
\label{fig:valmetrics}
\end{figure*}

The GRU model was selected as the final architecture primarily because of its exceptional calibration performance. In general, it showed strong predictive performance and clinical utility, as seen in Figure \ref{fig:valmetrics}. The model achieved an AUC-ROC of 0.89, indicating a high discriminative ability in distinguishing between mortality visits and non-event visits. Despite the class imbalance, given that 67\% of the patients in that cohort had passed away, the precision-recall curve yielded a PR-AUC of 0.871, which exceeds the positive prevalence at the baseline visit-level (0.46) for the 180‑day endpoint in the external cohort. To satisfy the operating-point protocol and the predefined $\geq 85\%$ sensitivity constraint in the external cohort, we set $\theta = 0.35$, which supports direct comparisons of alert burden and calibration under a fixed clinical safety requirement. Finally, the trade-off between PPV and NPV at this selected threshold reveals the model\textquotesingle{}s effectiveness in managing the balance between false alarms and missed events, ensuring reliable risk stratification for patient care.

\subsubsection{Clinical Impact Metrics}

A more rigorous evaluation of the GRU risk model reveals a practical balance between predictive foresight and clinical workload. The system generates 18.3 alerts per 100 patient-visits, corresponding to an alarm density where 18.31\% of all visits result in an active alert status, which poses a slight clinical burden on hospital staff. Crucially, the model demonstrates significant lead time for actionable intervention, with a median TIW of 151.0 days for true positive cases. In contrast, false positive alerts exhibit a much shorter noise duration, with a median TIW of 59.0 days.

\begin{figure}[h]
\vspace{-3mm}
     \centering
     \includegraphics[width=\linewidth]{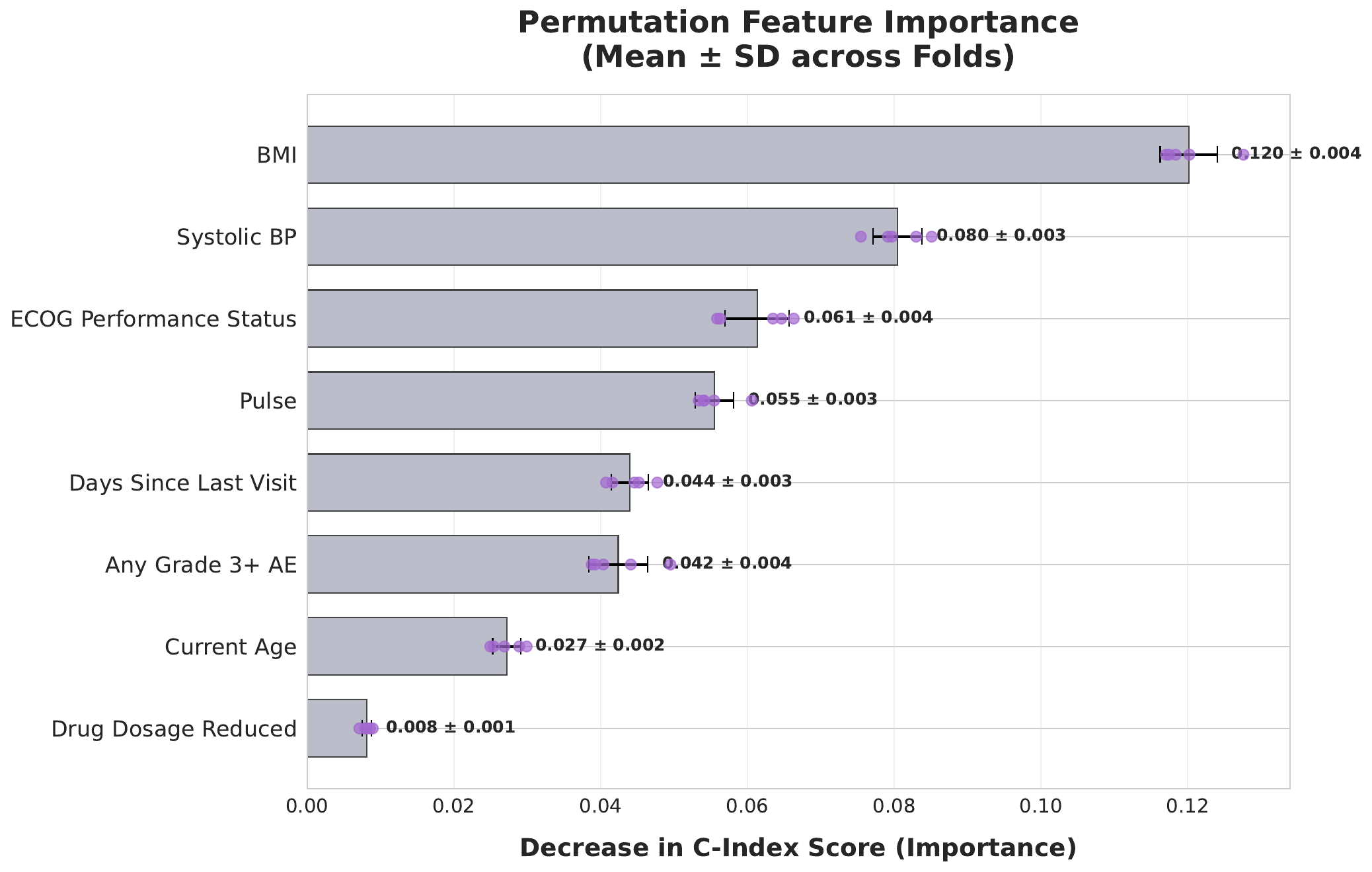}
     \caption{\footnotesize \textbf{Permutation-Based Feature Importance Ranking for the  GRU survival model on the Internal Validation Cohort.} To quantify importance, we measured the mean decrease in the Concordance Index (C-index) after permuting each feature across 3-fold stratified cross-validation. Horizontal bars show the mean decrease in C-index, and error bars show $\pm 1$ standard deviation (SD). The dots show fold-specific importance scores.\\}
     \label{fig:featureimp}
     \vspace{-7mm}
\end{figure}

The permutation feature importance analysis reveals that metabolic and cardiovascular metrics are the primary drivers of the GRU model\textquotesingle{}s predictive accuracy, with BMI and systolic blood pressure showing the most significant impact on C-index degradation at $0.120 \pm 0.004$ and $0.080 \pm 0.003$, respectively (Figure~\ref{fig:featureimp}). Clinical indicators of functional status and autonomic stability, specifically ECOG performance status ($0.061 \pm 0.004$) and pulse ($0.055 \pm 0.003$), also serve as critical inputs for mortality risk stratification. While temporal factors like days since last visit ($0.044 \pm 0.003$) and the occurrence of any grade 3+ adverse events ($0.042 \pm 0.004$) provide moderate predictive value, the model is least sensitive to current age and drug dosage reduced, which contribute minimally to overall performance. We interpret permutation importance as indicating model reliance on correlated markers rather than causal effects.
In mCRPC, reduced BMI and abnormal blood pressure can serve as inexpensive proxies for systemic decline that often precedes death.
This pattern can explain their high importance when the feature set excludes oncology-specific biomarkers. We also note that visit timing and missingness patterns can be informative in trial data. To quantify their contribution, our future sensitivity analyses plan will measure temporal utilization signals (e.g., days since the last visit).

\section{Discussion}

The results of this study demonstrate that a gated recurrent unit (GRU) architecture can effectively use longitudinal clinical trial data to predict 180-day mortality in mCRPC patients with high precision. Specifically, the GRU achieved a C-index of 0.87 and a balanced accuracy of 0.88 in the internal validation cohort. When subjected to external validation, the C-index was 0.81 and the balanced accuracy 0.80. The model maintained robust discriminative power with an AUC-ROC of 0.89 and a PR-AUC of 0.871, significantly outperforming the baseline event rate of 0.46. Beyond raw predictive accuracy, the GRU showed high clinical utility: it maintained a manageable clinical burden of 18.3 alerts per 100 patient-visits while providing a median lead time of 151.0 days for true positive alerts. This substantial lead time suggests that the model does not merely identify patients in immediate terminal decline but rather captures the physiological transition toward mortality months in advance.

A comparative analysis of the candidate architectures highlights a trade-off between raw discrimination and calibration. While both the GRU and Random Survival Forest (RSF) models achieved an identical C-index of 0.87 in the development cohort, their performance diverged slightly during external validation. The RSF model, despite showing high specificity (0.89) and PPV (0.87) in the validation set, suffered from substantial miscalibration, characterized by a slope of 1.34 and an intercept of -0.20. In contrast, the GRU model demonstrated superior reliability with an external calibration slope of 0.93 and an intercept of 0.07. The sequential nature of the GRU likely allowed it to better internalize the temporal dependencies of physiological markers, such as the identified primary drivers, BMI and systolic blood pressure, compared to the static landmarking approach of the RSF. By prioritizing calibration, the GRU ensures that the predicted probabilities correspond closely to actual observed risks, which is a prerequisite for trustworthy clinical decision support.

The clinical implications of these findings are profound, particularly for the management of aggressive mCRPC where treatment response is often heterogeneous. The ability to predict terminal progression with a median lead time of approximately five months provides a critical \textquotesingle{}window of opportunity\textquotesingle{} for clinicians and patients. In practice, a high-risk alert could trigger a multi-step clinical response: first, a reassessment of aggressive treatment regimens that may carry a high toxicity burden with diminishing returns, and second, the early introduction of palliative care and end-of-life support. Furthermore, since the model relies on routine clinical markers like BMI, blood pressure, and ECOG status, it could be integrated into existing electronic health record systems without requiring expensive or invasive new biomarkers. This framework shifts the prognostic paradigm from reactive crisis management to proactive, patient-centered planning, potentially improving the quality of life for patients in their final months.

\subsection{Limitations and Future Work}

Several limitations warrant consideration. First, the use of clinical trial data may limit generalizability to real-world populations, a concern compounded by the predominantly White cohort (82.3\% and 90.6\%). Given limited racial diversity and restrictive inclusion criteria, we cannot claim equitable performance across demographic groups. As part of our future plan, we will evaluate subgroup calibration and alert burden (e.g., alarm fatigue risk) and validate across multiple sites in underrepresented populations.
 Second, the model relies solely on structured data, omitting potentially informative unstructured clinical notes. Third, because $\theta$ is chosen to meet a sensitivity constraint within each evaluation dataset, specificity and PPV reflect an achievable clinical operating point. In deployment, we would estimate $\theta$ on a small local calibration set and then fix it prospectively. Finally, while the external validation cohort only included the control arm for that clinical trial, the 1:1 baseline randomization of participants suggests that this selection is unlikely to introduce significant systematic bias into the reported performance metrics. 

To reduce performance gaps without centralizing sensitive data, we propose a Diversity Validation Plan that tests subgroup calibration and alert burden parity across geographically dispersed healthcare systems. Moreover, we mitigated label ambiguity by restricting evaluation to visits with observable 180-day outcomes, and we plan to adopt discrete-time survival training based on censoring to keep censored visits and further improve transportability to real-world EHR settings with heterogeneous follow-up. We also aim to extend this methodology to other cancer subtypes and employ privacy-preserving technologies, such as Federated Learning, to develop equitable models across institutions.

\section{Conclusion}
This study developed and externally validated a longitudinal GRU-based model that estimates 180-day mortality risk in mCRPC from standard clinical visit data. The GRU\textquotesingle{}s calibration (slope 0.93)  makes it a reliable choice for clinical implementation, providing trustworthy risk scores that align with actual patient outcomes. With a median lead time of 151 days and a sustainable alert density of 18.3 per 100 visits, the model offers a practical solution for enhancing end-of-life care and personalizing treatment strategies. Future work should focus on integrating these longitudinal alerts into electronic health records to evaluate their impact on clinical outcomes in real-world settings.

%


\bibliographystyle{IEEEtran}
\bibliography{References}

\end{document}